\documentclass[review]{elsarticle}

\usepackage{lineno,hyperref}
\modulolinenumbers[5]

\usepackage{epsfig}
\usepackage{graphicx}
\usepackage{diagbox}
\usepackage{amsmath,amssymb,amsfonts,bm}
\usepackage{graphicx}
\usepackage{multirow}

\usepackage{algorithm}
\usepackage{algorithmicx}
\usepackage[noend]{algpseudocode}
\usepackage[caption=false]{subfig}

\journal{Journal of \LaTeX\ Templates}

\begin{document}

\begin{frontmatter}

\title{Evaluation of Inference Attack Models for \\ Deep Learning on Medical Data}


\author[gdut]{Maoqiang Wu}
\author[uh]{Xinyue Zhang}
\author[uh]{Jiahao Ding}
\author[uh]{Hien Nguyen}
\author[gdut]{Rong Yu}
\author[uh]{Miao Pan}
\author[meth]{Stephen T. Wong}
\address[gdut]{School of Automation, Guangdong University of Technology, Guangzhou 510006, China}
\address[uh]{Electrical and Computer Engineering Department, University of Houston, TX 77004, USA}
\address[meth]{Systems Medicine and Bioengineering, Houston Methodist Cancer Center, 6445 Main Street, TX 77030, USA}




\begin{abstract}
Deep learning has attracted broad interest in healthcare and medical communities. However, there has been little research into the privacy issues created by deep networks trained for medical applications. Recently developed inference attack algorithms indicate that images and text records can be reconstructed by malicious parties that have the ability to query deep networks. This gives rise to the concern that medical images and electronic health records containing sensitive patient information are vulnerable to these attacks. This paper aims to attract interest from researchers in the medical deep learning community to this important problem. We evaluate two prominent inference attack models, namely, attribute inference attack and model inversion attack. We show that they can reconstruct real-world medical images and clinical reports with high fidelity. We then investigate how to protect patients' privacy using defense mechanisms, such as label perturbation and model perturbation. We provide a comparison of attack results between the original and the medical deep learning models with defenses. The experimental evaluations show that our proposed defense approaches can effectively reduce the potential privacy leakage of medical deep learning from the inference attacks.
\end{abstract}

\begin{keyword}
attribute inference attack, model inversion attack, medical machine learning, collaborative inference
\end{keyword}

\end{frontmatter}



\section{Introduction}

Deep learning has become increasingly popular in healthcare and medicine areas. Medical institutions hold various modalities of medical data, such as electronic health records, biomedical images, and pathology test results. Based on medical data, deep neural network models are trained to address necessary healthcare concerns \cite{miotto2018deep}. Examples include but not limited to: 1) deep learning models based on medical records outperformed traditional clinical models for detecting patterns in health trends and risk factors \cite{rajkomar2018scalable}; 2) deep learning model had high sensitivity and specificity for detecting diabetic retinopathy and macular edema in retinal fundus photographs \cite{gulshan2016development}; 3) a mammography-based deep learning model was more accurate than traditional clinical models for predicting breast cancer risk \cite{yala2019deep}.

However, the application of deep learning in healthcare is confronted with privacy threats. The medical records contain personal private information like drug usage patterns of the individual patient. Medical institutions also hold patients' profile information such as home address, gender, age, etc. The private information might be unwittingly leaked when the aforementioned data is used for training a deep learning model \cite{fredrikson2015model,shokri2017membership}. For example, attribute inference attacks \cite{fredrikson2015model} can utilize the trained model and incomplete information about a data point to infer the missing information for that point. The adversary could exploit such an attack to infer the target private information according to partial information on medical records. Another instance is model inversion attacks that enable the image data used for classification inference to be recovered based on the intermediate output of convolutional neural networks (CNNs) \cite{he2019model}. The adversary could deploy model inversion attack to recover the medical images of any target patient and infer private health conditions accordingly.

The risk of privacy leakage makes medical institutions increasingly less willing to share their data. This inevitably slows down the research progress at the intersection of deep learning and healthcare. Thus, it is necessary to evaluate the potential hazards of various attack models on medical data and develop corresponding defenses on such attacks.

\begin{figure*}[!htb]
   \centering
   \subfloat[Attribute inference attack]{\label{subfig:AIA}
      \includegraphics[width=.75\textwidth]{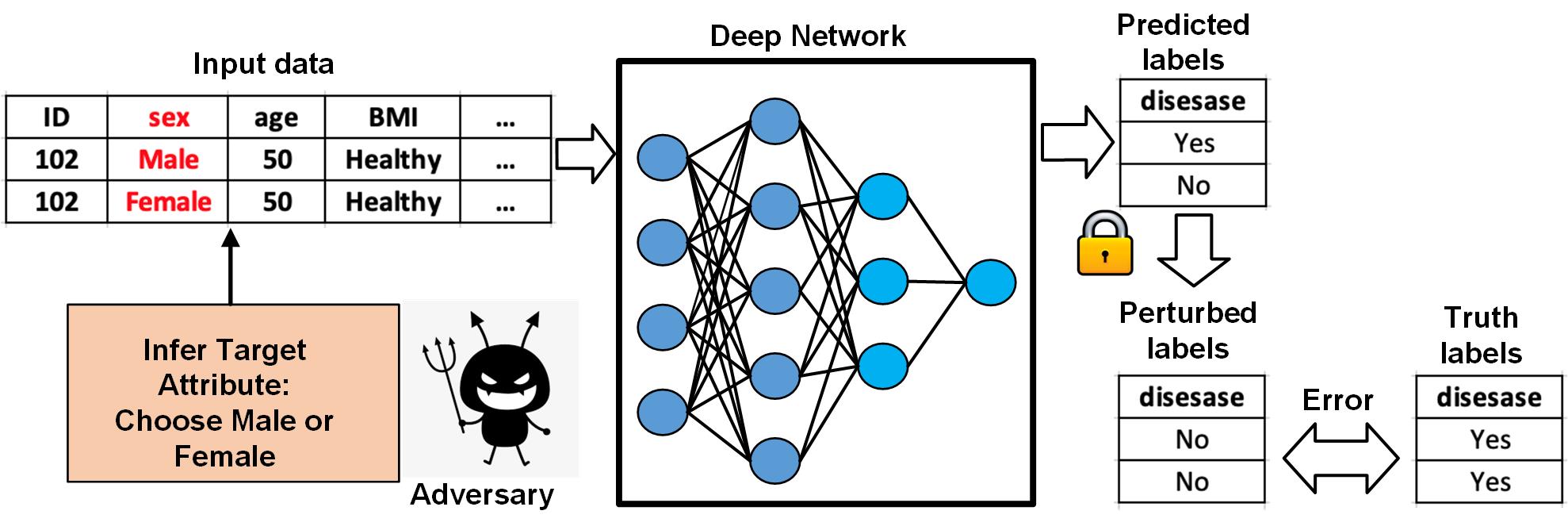}}\\
   \subfloat[Model inversion attack]{\label{subfig:MIA}
      \includegraphics[width=.75\textwidth]{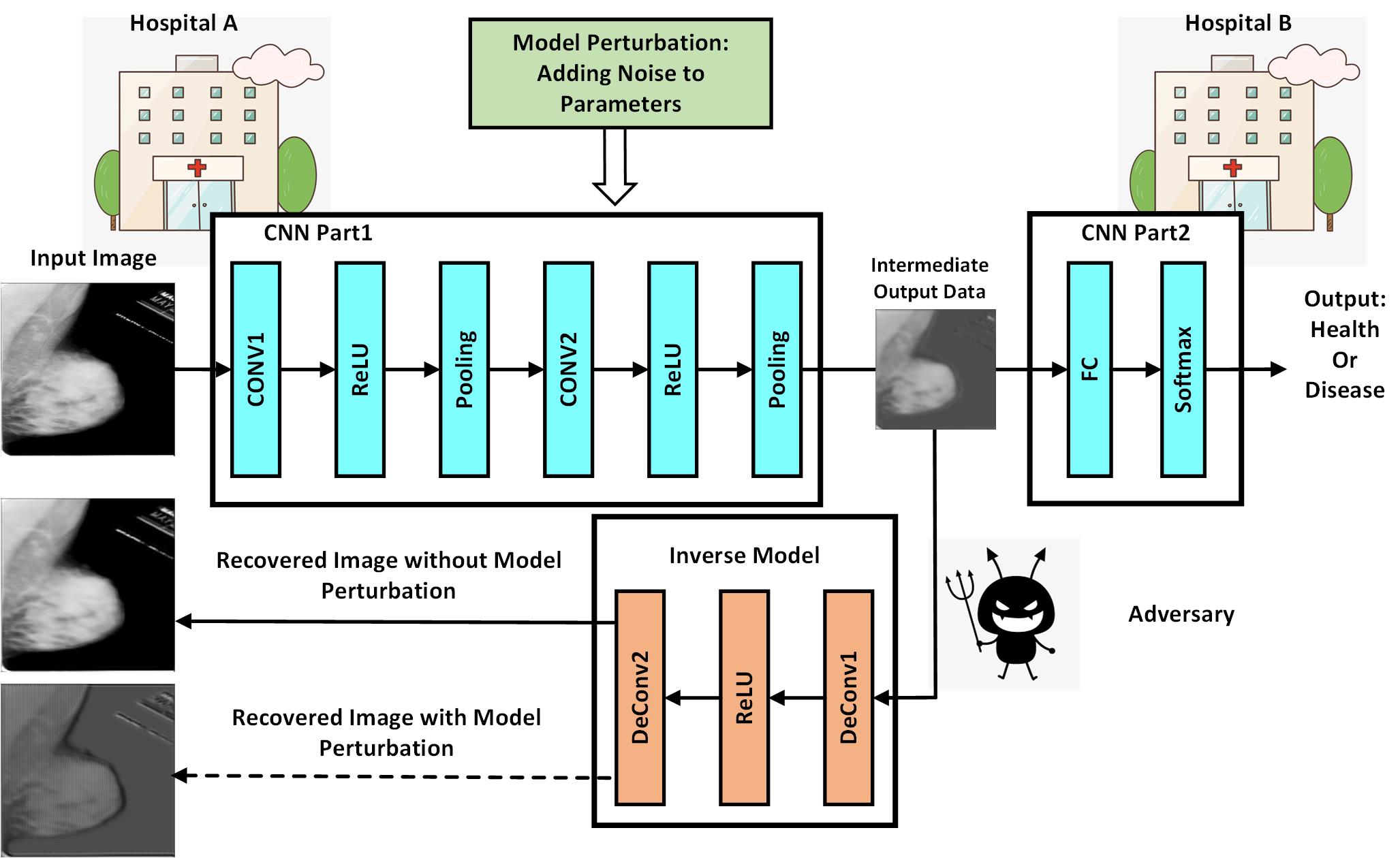}}
   \caption{Inference attack models and defense approaches for medical deep learning.}\label{fig:System}
\end{figure*}

In this paper, we attempt to implement two types of attack models on medical data, which are shown in Fig. \ref{fig:System}. We use attribute inference attack to infer the sensitive attributes in medical record data according to the rest attributes and class labels, when training a deep learning model. We also employ model inversion attack to recover the medical image data based on the intermediate inference output. Against those attacks, we present two types of inference attack defense mechanisms. Label perturbation provides a way to add noise into confidence scores and thus hinders the privacy leakage from model prediction. Model perturbation is proposed to add noise into parameters of deep network models and thus disturb the privacy disclosure in model inference. Experimental results show that both attacks successfully disclose the private medical information used in training and inference processes, and the attacks are not effective any more under the proposed inference attack defense mechanisms.

The main contributions of this paper are summarized as follows.
\begin{itemize}
    \item We evaluate attribute inference attack and model inversion attack on medical data. That demonstrates the privacy vulnerability of deep learning models, which limits their applications in the medical area. As far as we know, we are the first to evaluate the model inversion attack on medical data.
    \item We present inference attack defenses based on label perturbation and model perturbation. The mechanisms can significantly alleviate the privacy breaches of medical data in both the training phase and the inference phase.
\end{itemize}

The rest of this paper is organized as follows. We summarize the related work in Section 2. In Section 3, we describe the attribute inference attack and model inversion attack. We propose label perturbation mechanism and model perturbation mechanism against the two attacks in Section 4. Simulation results are presented in Section 5. Then we conclude the paper in Section 6.


\section{Related Work}

There are different types of privacy attacks against training and inference data. These attacks severely threaten patients' privacy when deep learning is used in the healthcare area. The first type is membership inference attacks \cite{nasr2018machine,shokri2017membership}, which tries to infer whether a target sample is contained in the dataset. The second type is model encoding attacks \cite{song2017machine}, the adversary who directly accesses to the training data can encode the sensitive data into the trained model and then retrieve the encoded sensitive information. The third type is attribute inference attack, given some attributes of the dataset, the adversary could infer the sensitive attribute. The fourth type is model inversion attack, given a deep learning model and some features of input data, the adversary could recover the rest of the features of the input data. In this paper, we choose two prominent inference attacks, namely attribute inference attack and model inversion attack, which may reconstruct medical images and clinical reports and be more threatening to patients' privacy. We evaluate their attack performance on medical records and medical images, and then propose defense methods against these two inference attacks.

{\bf{Attribute Inference Attack}} Attribute inference attack is studied in various areas. Gong et al. \cite{gong2016you,gong2018attribute} studied attribute inference attacks to infer the users' sensitive attribute of social networks by integrating social friends and behavioral records. May et al. \cite{mei2018image} proposed a new framework for inference attacks in social networks, which smartly integrates and modifies the existing state-of-the-art CNN models. Qian et al. \cite{qian2016anonymizing} demonstrated that knowledge graphs can strengthen de-anonymization and attribute inference attacks, and thus increase the risk of privacy disclosure. However, few studies evaluate the attribute inference attacks in the healthcare area. In this paper, we adopt the same attribute inference attack in \cite{fredrikson2014privacy} which infers the sensitive attributes based on confident cores in predictions and is conveniently deployed on healthcare data. We propose a label perturbation method to effectually defend against the attribute inference attack.

{\bf{Model inversion attack}} Model inversion attack is an outstanding attack to recover the input data of deep neuron networks. He et al. \cite{fredrikson2015model} proposed a model inversion attack to recover input images via the confidence score generated in the softmax model. He et al. \cite{he2019model} proposed a model inversion attack to reconstruct input images via the intermediate outputs of the neural network. Hitaj et al. \cite{hitaj2017deep} utilized Generative adversarial network (GAN) to recover the image in a collaborative training system. In this paper, we adopt the same model inversion attack in \cite{he2019model} by considering the medical collaborative deep learning scenario, where two hospitals hold different parts of a deep neuron network and collaborate to complete the training and inference via transmitting the intermediate output information.  As far as we know, we are the first to evaluate the model inversion attack on medical data via intermediate output information. We propose an effective and convenient perturbation method instead of using the defenses suggested in \cite{he2019model}, i.e., combining Trust Execution Environment and Homomorphic Encryption that requires special architecture support and huge computational burden.

{\bf{Other Attacks against Machine Learning}}
Besides attribute inference attack and membership inference attack, there exist numerous other types of attacks against ML models \cite{247644,ganju2018property,gu2017badnets,guo2018lemna,jagielski2020high,ji2018model,leino2019stolen,li2020membership,li2019prove,ling2019deepsec,Trojannn,oh2019towards,papernot2017practical,papernot2016limitations,papernot2018sok,quiring2019misleading,salem2020dynamic,shafahi2018poison,she2020neutaint,tramer2018ensemble,tramer2016stealing,wang2018stealing,wang2019neural}. A major attack type is adversarial examples  \cite{papernot2018sok,papernot2017practical,papernot2016limitations,tramer2018ensemble}. In this setting, the adversary tries to carefully craft noise and add them to the data samples aiming to mislead the target classifying. In addition, a similar type of attack is backdoor attack, where the adversary tries to embed a trigger into a trained model and to exploit when the model is deployed \cite{gu2017badnets,Trojannn,salem2020dynamic}. Another line of work is model stealing attack, \cite{tramer2016stealing} proposed the first attack on inferring a model’s parameters and the related works focus on inferring a model’s hyperparameters \cite{oh2019towards,wang2018stealing}.

{\bf{Possible Defenses}} To defend against the privacy attack, many researchers focused on defense methods. Trust Execution Environment \cite{sabt2015trusted} is specialized hardware for secure remote computation and data confidentiality protection against privileged adversaries. Homomorphic Encryption \cite{acar2018survey} allows the training and inference operations on encrypted input data, so the sensitive information will not be leaked. However, these methods require special architecture support and a huge computational burden. Differential Privacy (DP) \cite{ding2019stochastic} adds noise into the training model and there exists a trade-off between usability and privacy. However, our attacks mainly focus on the inference phase rather than the training phase and thus the DP methods are not suitable to defend against our attacks. We propose label perturbation that adds noise in the predicted label to defend attribute inference attack and mode perturbation that adds noise into the after-trained model to defend model inversion attack. The proposed methods are effective and convenient for application. We also give the results of the trade-off between model accuracy and attack performance. These results provide an intuitive guide for medical staff to adjust the defenses against the two inference attacks.

\section{Inference Attack Models}

\subsection{Vulnerability of Medical Deep Learning}
Via the use of deep learning algorithms, medical institutions can improve the rate of correct diagnosis
\cite{rajkomar2018scalable}. In the training phase, deep neural networks are trained based on input medical data and output diagnosis results to learn the inherent relationships between them. In the inference phase, the after-trained deep neural networks can achieve high-accurate diagnosis results given new medical data as input. However, during the training phase and inference phase, the adversary could adopt attack methods to infer or recover the input medical data which contains sensitive information of patients. In this paper, we evaluate two prominent inference attacks, i.e., attribute inference attack and model inversion attack. As for attribute inference attack, we assume that the adversary knows the attributes of all the patients except the sensitive attributes when the patients' medical records are used as input. This assumption applies to the cases that the attacker can search the rest attributes such as age and gender from the public database. The adversary utilizes the inherent relationships among attributes and labels to recover the patients' sensitive attributes of input medical records. As for model inversion attack, we take the vulnerability of collaborative deep learning as an example, which provides an efficient paradigm to accelerate the learning and prediction process. The fundamental idea is to split a deep neural network into two parts. For example as Fig. \ref{subfig:MIA}, in medical collaborative learning, the first few layers are stored in Hospital A while the rest are kept in Hospital B. In the collaborative training mode \cite{vepakomma2018split}, Hospital A sends the outputs of the cut layer to Hospital B and then retrieve the gradients of the cut layer. In the collaborative inference mode \cite{he2019model,li2019edge}, Hospital A sends the outputs of the cut layer to Hospital B and retrieves the final results. The model training and inference processes are collaboratively carried out without sharing the raw data. However, the shared intermediate output information may be leaked during the transmission. Given the information, the adversary could recover the raw data with model inversion attack and thus compromise the data privacy of Hospital A.

\subsection{Attribute Inference Attack}\label{subsec:attr_attack}


As shown in Fig. \ref{subfig:AIA}, attribute inference attack \cite{fredrikson2014privacy} enables an adversary to deduce sensitive attributes in patients' medical records. In this setting, the goal of the adversary is to guess the value of the sensitive features of a data point, e.g., sex attribute, given public knowledge about the data sample and the access to the model. Let $(x, y)$ denote a data sample where $x$ denotes the input patient information, and $y$ is the label of this data sample. We assume that a deep network $f(x)$ takes the input $x$ to predict the output $y$. The network's parameters are optimized by reducing the discrepancy between the predicted value $f(x)$ and the true outcome $y$ measured by the cross-entropy loss. We assume there are $d$ attributes in a data sample $x$ and let $x_d$ be a sensitive attribute in $x$ that an attacker wants to learn. Given the values of attributes $x_1, x_2, \cdots, x_{d-1}$, the prior probabilities of all attributes and the access to the model $f(x)$, the attacker aims to find the value of $x_d$ to maximize the posterior probability $P(x_d|x_1, x_2, \cdots, x_{d-1}, f(x))$. Therefore, the attacker can obtain the value of the sensitive attribute $x_d$.



\subsection{Model Inversion Attack}

As shown in Fig. \ref{subfig:MIA}, model inversion attack \cite{he2019model} enables an adversary to recover an input medical image $x_0$ from the corresponding intermediate output $v_0 = f_{\theta}(x_0)$, where $f_{\theta}$ is the former layers of the model in Hospital A. We consider the black box attack setting, where the adversary does not know the structure or parameters $\theta$ of $f_{\theta}$ but he could query the black-box model, i.e., he could input the arbitrary data $X$ into the model and observe the intermediate outputs $f_{\theta}(X)$. This assumption happens to the use case where Hospital A releases its APIs to other medical entities as training and inference services. In this setting, we build an inverse network model that learns the inverse mapping from output to input without the original model information. Roughly, the inverse model $g_{\omega} \approx f^{-1}_{\theta}$ can be regarded as the approximated inverse function of $f_{\theta}$, where $v=f_{\theta}(x)$ is input and $x$ is output.

Algorithm \ref{Algorithm1} shows the detailed model inversion attack consisting of four phases. In the observation phase, the adversary uses a cluster of samples $X={x_1,\cdots,x_n}$ as inputs to query $f_{\theta}$ and obtain $V={f_{\theta}(x_1),\cdots,f_{\theta}(x_n)}$. Here the sample set $X$ is assumed to follow the same distribution of $x_0$. The assumption applies to the case that the radiologic images usually follows the same distribution. In the training phase, the adversary trains the inverse network $g_{\omega}$ by using $V$ as inputs and $X$ as targets. We exploit the $l_2$ norm in the pixel space as the loss function, which is given as
\begin{equation}
    l(\omega;X)=\frac{1}{n}{\sum^{n}_{i=1}}\left\|g_{\omega}\big(f_{\theta}( x_i ) \big)-x_i\right\|^2_2.
\end{equation}
In particular, the structure of $g_{\omega}$ is not necessarily related to $f_{\theta}$. In our experiment, an entirely different architecture is leveraged for the attack. In the recovery phase, the adversary leverages the trained inverse model to recover the raw data from the intermediate value: $x_{0}'= g_{\omega}(v_0)$.
\begin{algorithm}[!htb]
\caption{Model Inversion Attack Algorithm}
\label{Algorithm1}
\hspace*{0.02in} {\bf Input:}
input data $X={x_1,x_2,\cdots,x_n}$ of the same distribution from target data $x_0$, output $v_0$ of target data, batch size $B$, epoch number $E$, learning rate $\eta$\\
\hspace*{0.02in} {\bf Output:}
recovered data $x_{0}'$
\begin{algorithmic}[1]
\State query the model by input data $V=f_{\theta}(X)$
\State initialize  $\omega_0$
\For{each epoch $ t \in T$}
\State $\beta \leftarrow$ (split $V$ into batches of size $B$)
\For{each batch $b \in \beta$}
\State $\omega_{t+1} \leftarrow \omega_t - \eta \nabla l(\omega_t;b) $
\EndFor
\EndFor
\State recover the target data $x_{0}'= g_{\omega}(v_0)$
\State \Return $x_{0}'$
\end{algorithmic}
\end{algorithm}


\section{Inference Attack Defense Mechanisms}

\subsection{Label Perturbation Based Protection}
We apply randomized responses~\cite{warner1965LocalPrivacy} to protect the learning model output labels of each data sample against attribute attacks. Intuitively, given a flipping probability $p$, for a binary classification, the predicted label $y$ will be flipped with $p$. Similarly, we assume the class set $\mathcal{C} = \{1,2,\cdots, C\} (C>2)$, the predicted label $y \in \mathcal{C}$ will be perturbed with the probability of $p$. If the predicted label $y$ is going to be replaced, there is $1/(C-1)$ probability for each one of the other $C-1$ classes that the original label $y$ will be substituted by the corresponding class. The inference accuracy of the attribute inference attack will deteriorate when the adversary obtains the inaccurate predicted label. Although the training performance can be influenced by the label perturbation, by controlling the flipping probability $p$ carefully, we can still have an acceptable training model.


\subsection{Model Perturbation Based Protection}

To defend against model inversion attack, we adopt model perturbation in CNN model. Different from the label perturbation that adds noise into predicted label, model perturbation adds noise into model parameters $\theta$ (weights and bias) before the forward propagation is implemented. Specifically, we use Gaussian mechanism with expectation $0$ and variance $\sigma$ to generate noise and add it into model parameters, which is given as
\begin{equation}
    \theta = \theta + \mathcal{N}(0, \sigma^2 I).
\end{equation}
Accordingly, the output of the cut layer is perturbed in collaborative deep learning. Model inversion attack becomes difficult to build an accurate mapping from the output to the input image and thus the image recovery quality decreases.


\section{Performance Evaluation}


\subsection{Attribute Inference Attack}

\subsubsection{Experiment Settings}

We evaluate attribute inference attack and label perturbation approach on two public medical record datasets: cardiovascular disease dataset and heart disease dataset. The cardiovascular disease dataset consists of $70,000$ records, $11$ feature attributes including sensitive information such as age and gender, and labels indicating the presence or absence of cardiovascular disease. Heart disease dataset~\cite{detrano1989international} contains $13$ attributes, $303$ instances, and labels referring to the presence of heart disease in individual patients. We split the dataset into training set and testing set with 80\% and 20\%. Our experiments use a multilayer perceptron classifier including 2 hidden layers with 100 neurons in each hidden layer.

\begin{figure}[!htb]\centering
\vspace{-.2in}
   \subfloat[Smoking]{\label{subfig:decay}
      \includegraphics[width=.45\textwidth]{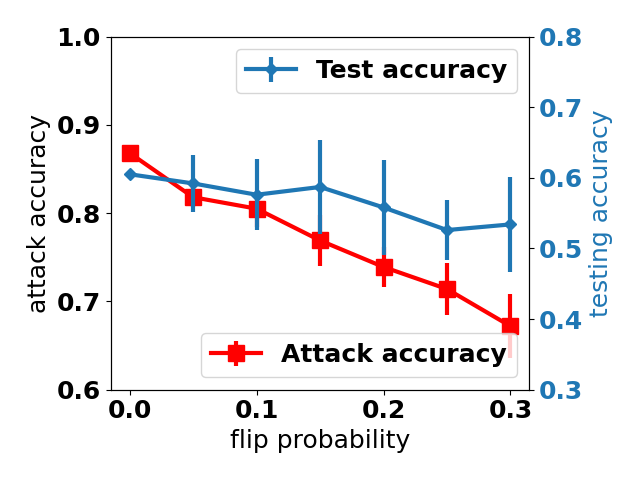}}
   \subfloat[Alcohol intake]{\label{subfig:noise}
      \includegraphics[width=.45\textwidth]{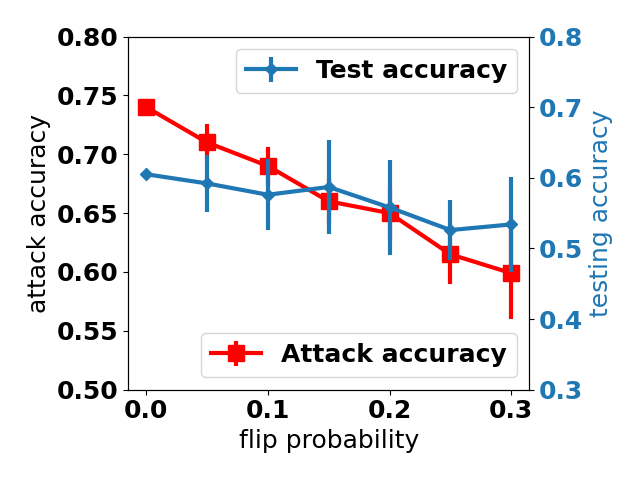}}
   \caption{Attribute attack performance on the ``cardiovascular disease" dataset.}\label{fig:attr_heart}
\end{figure}
\vspace{-.2in}

\subsubsection{Evaluation Results}
\begin{figure}[!htb]\centering
\vspace{-.2in}
   \subfloat[Smoking]{\label{subfig:decay}
      \includegraphics[width=.45\textwidth]{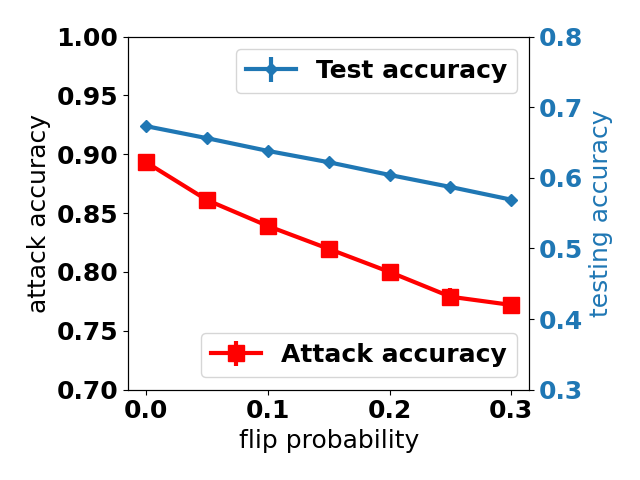}}
   \subfloat[Alcohol intake]{\label{subfig:noise}
      \includegraphics[width=.45\textwidth]{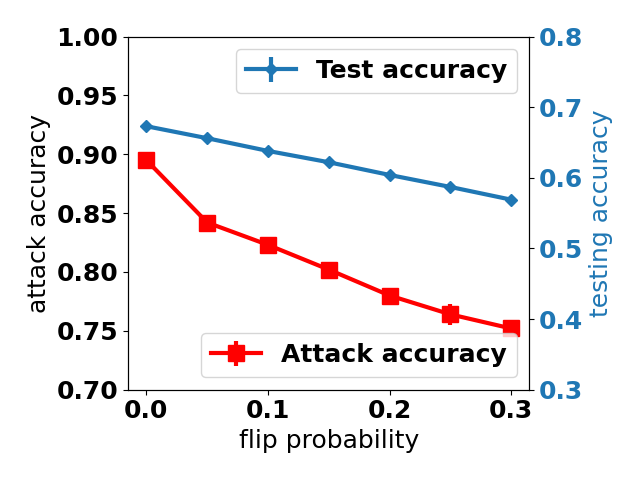}}
   \caption{Attribute attack performance on the ``cardiovascular disease" dataset.}\label{fig:attr_cardio}
\end{figure}

As described in Section~\ref{subsec:attr_attack}, in the experiments, we assume the attacker can obtain other information of a patient except for only one attribute and the marginal prior knowledge of the targeted attribute. We implement each experiment 10 times and show the mean value as the curve and the standard deviation as the error bar. The flip probability denotes the defense level. The higher flip probability means better defense. When the flip probability is equal to 0, it means no defense mechanism is applied. Fig. \ref{fig:attr_heart} and \ref{fig:attr_cardio} demonstrates the attack and defense performance on two datasets. We select two attributes from the ``heart" dataset, fasting blood sugar and gender, as the attacker's targets. For the ``cardiovascular" dataset, we choose smoking and alcohol intake as the target attribute. We can observe that the attack accuracy reduces with higher flip probability. Also, the testing accuracy degrades slightly, if the defense level is high.
\vspace{-.2in}

\subsection{Model Inversion Attack}

\subsubsection{Experiment Settings}

We evaluate model inversion attack and model perturbation-based defence on two public mammography datasets: MIAS \cite{suckling1994mammographic} and CBIS-DDSM \cite{lee2017curated}. All the images of MIAS dataset have been padded/clipped to $1024\times1024$. A total of $280$ samples are obtained from MIAS for training ($181$ normal, $57$ benign, $42$ malignant) while $50$ samples are used for testing ($26$ normal, $12$ benign, $12$ malignant). We clip and compress all the images of CBIS-DDSM to $256\times256$. A total of $2,326$ samples are obtained from CBIS-DDSM for training ($1,263$ benign, $1,063$ malignant) while $772$ samples are used for testing ($419$ benign, $353$ malignant). We adopt an CNN with $6$ convolution layers and $2$ fully connection layers on the two datasets. Each convolution layer has $32$ channels and kernel size is $3$. There is a maxpool layer after every two convolution layers. The model is split at 2nd, 4th, and 6th convolution layers. We select ADAM as our optimizer and set the learning rate as $0.001$.

\begin{figure*}[!htb]
  \begin{center}
    \includegraphics[width=0.7\textwidth,height=6.5cm]{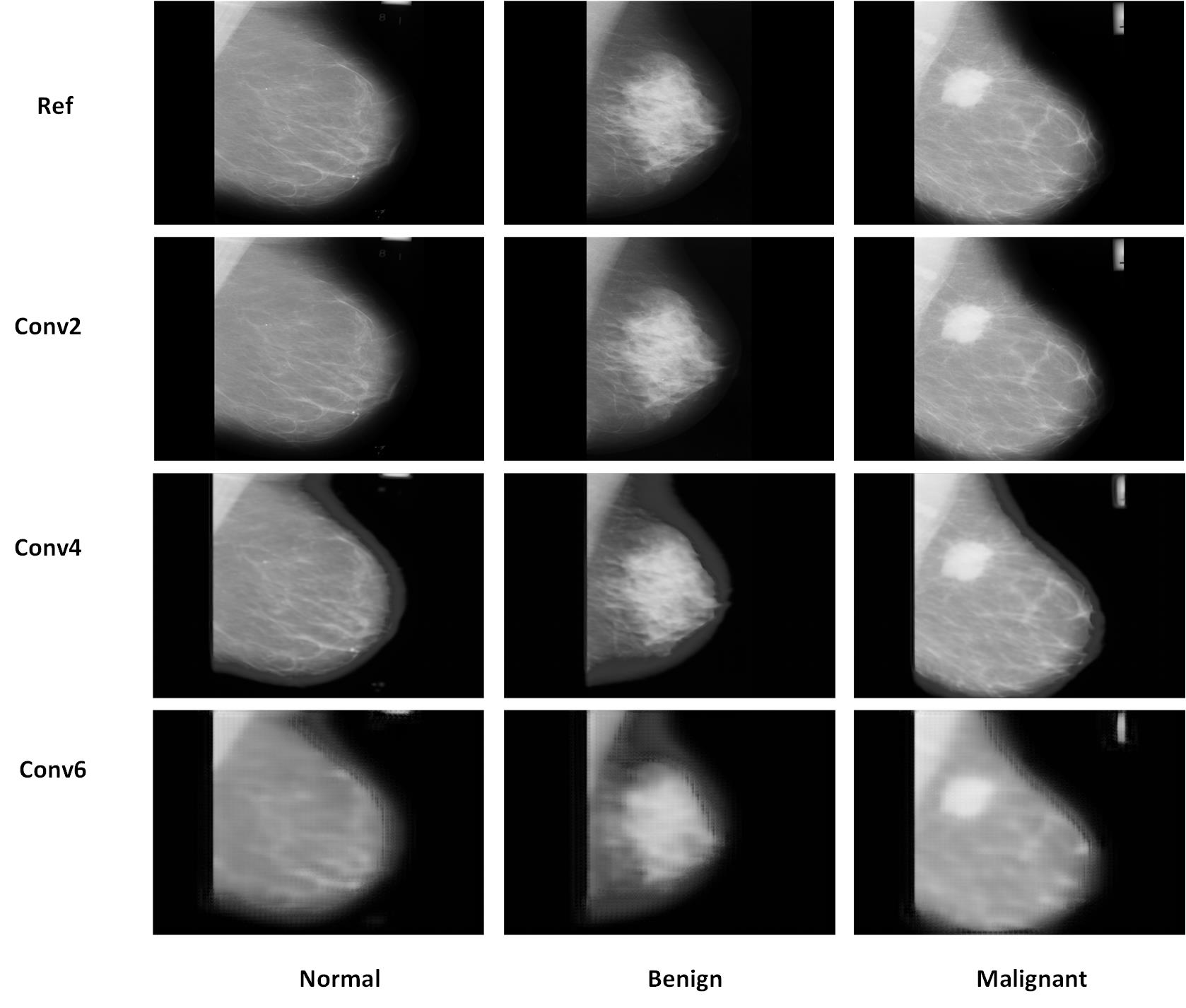}
  \end{center}
  \vspace{-5mm}
    \caption{Recovered MIA inputs via model inversion attack.}
    \label{fig:MIA_Performance_MIAS}
\end{figure*}

\begin{figure*}[!htb]
  \begin{center}
    \includegraphics[width=0.7\textwidth,height=6.5cm]{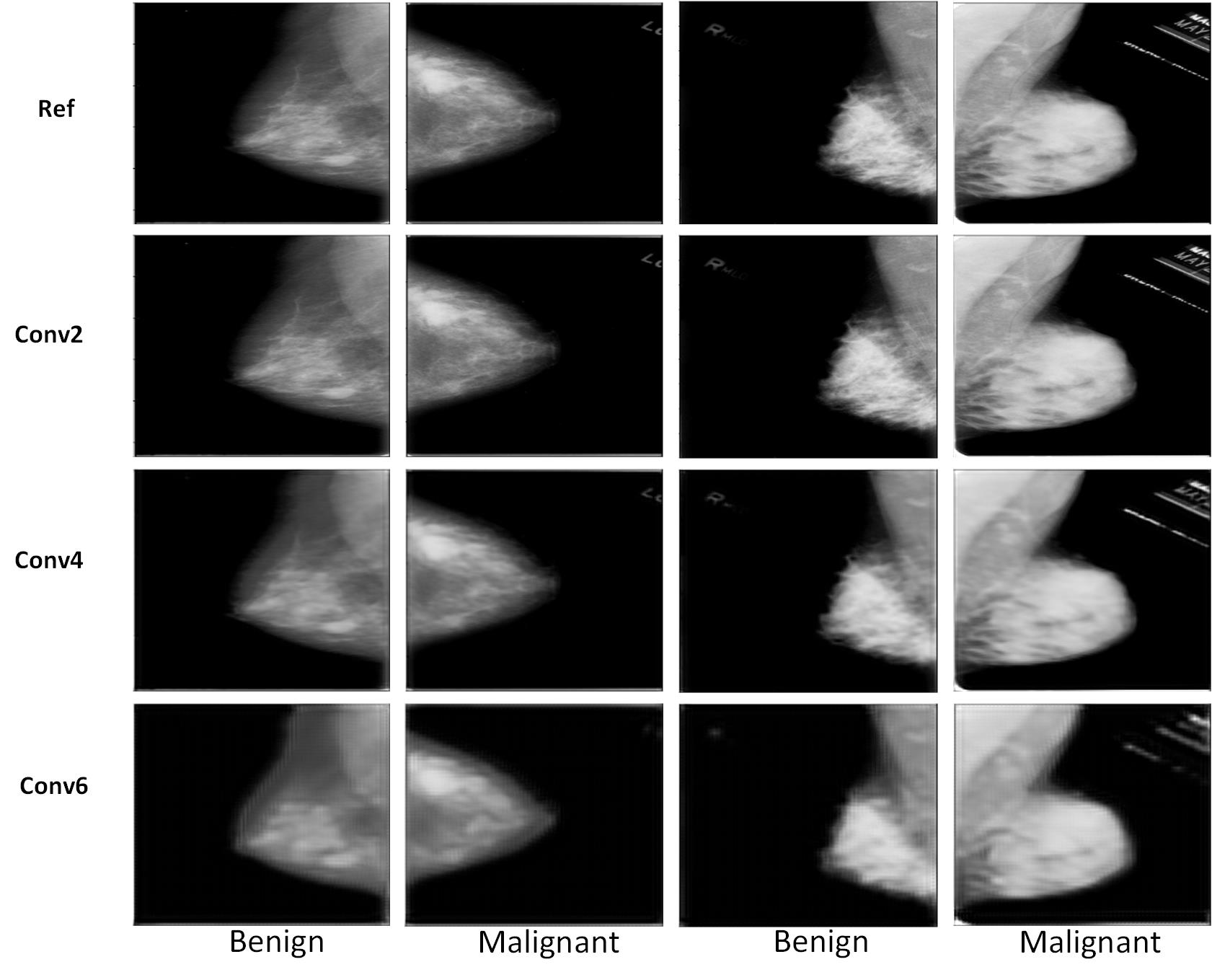}
  \end{center}
  \vspace{-5mm}
    \caption{Recovered CBIS-DDSM inputs via model inversion attack.}
    \label{fig:MIA_Performance_DDSM}
\end{figure*}

\begin{table*}[!htb]
	\renewcommand{\arraystretch}{1}
    \footnotesize
	\caption{MSE, PSNR, SSIM for model inversion attack with different split layers} \centering \tabcolsep=5pt
	\label{table1}
	\begin{tabular}{|p{1cm}|p{1cm}|p{1cm}|p{1cm}|p{1cm}|p{1cm}|p{1cm}|}
		\hline
		\multirow{2}*{} & \multicolumn{3}{c|}{MIAS} &  \multicolumn{3}{c|}{DDSM}\\
		\cline{2-7}
		\multirow{2}*{} & layer 2 & layer 4& layer 6 & layer 2 & layer 4& layer 6 \\
		\hline
		MSE & $2.925$ & $55.042$ & $88.839$ & $1.672$ & $29.649$ & $110.460$\\
      	\hline
		PSNR & $44.162$ & $31.039$ & $28.995$ & $46.385$ & $33.962$ & $28.269$\\
      	\hline
		SSIM & $0.999$ & $0.994$ & $0.990$ & $0.999$ & $0.995$ & $0.984$\\
        \hline
	\end{tabular}
\end{table*}

\subsubsection{Evaluation Results}
Fig. \ref{fig:MIA_Performance_MIAS} and Fig. \ref{fig:MIA_Performance_DDSM} show the recovered images via model inversion attack. When the split point is in a lower layer, the recovered images have high quality. When the split point is in a deeper layer, the recovered images become relatively blurry and lose certain details. But even if in the recovered image from the output of the 6th layer, the details within the breasts can still be clearly identified. The attackers could diagnose the patient's breast health given the recovered mammography image with the help of classification models or radiologists.

To quantify the attack results, we adopt three metrics, Mean-Square Error (MSE), Peak Signal-to-Noise Ratio (PSNR) and Structural Similarity Index (SSIM) \cite{hore2010image}, which are shown in Table \ref{table1}. MSE reflects pixel-wise similarity while PSNR measures the pixel-level recovery quality of the image. SSIM measures the human perceptual similarity of two images by considering their luminance, contrast, and structure. It ranges from $[0,1]$, where $1$ represents the most similar. When the split point is in a deeper layer, the recovered inputs have higher MSE, PSNR, and lower SSIM, which means the attack becomes harder.

\begin{figure*}[!htb]
  \begin{center}
    \includegraphics[width=0.7\textwidth, height=6.5cm]{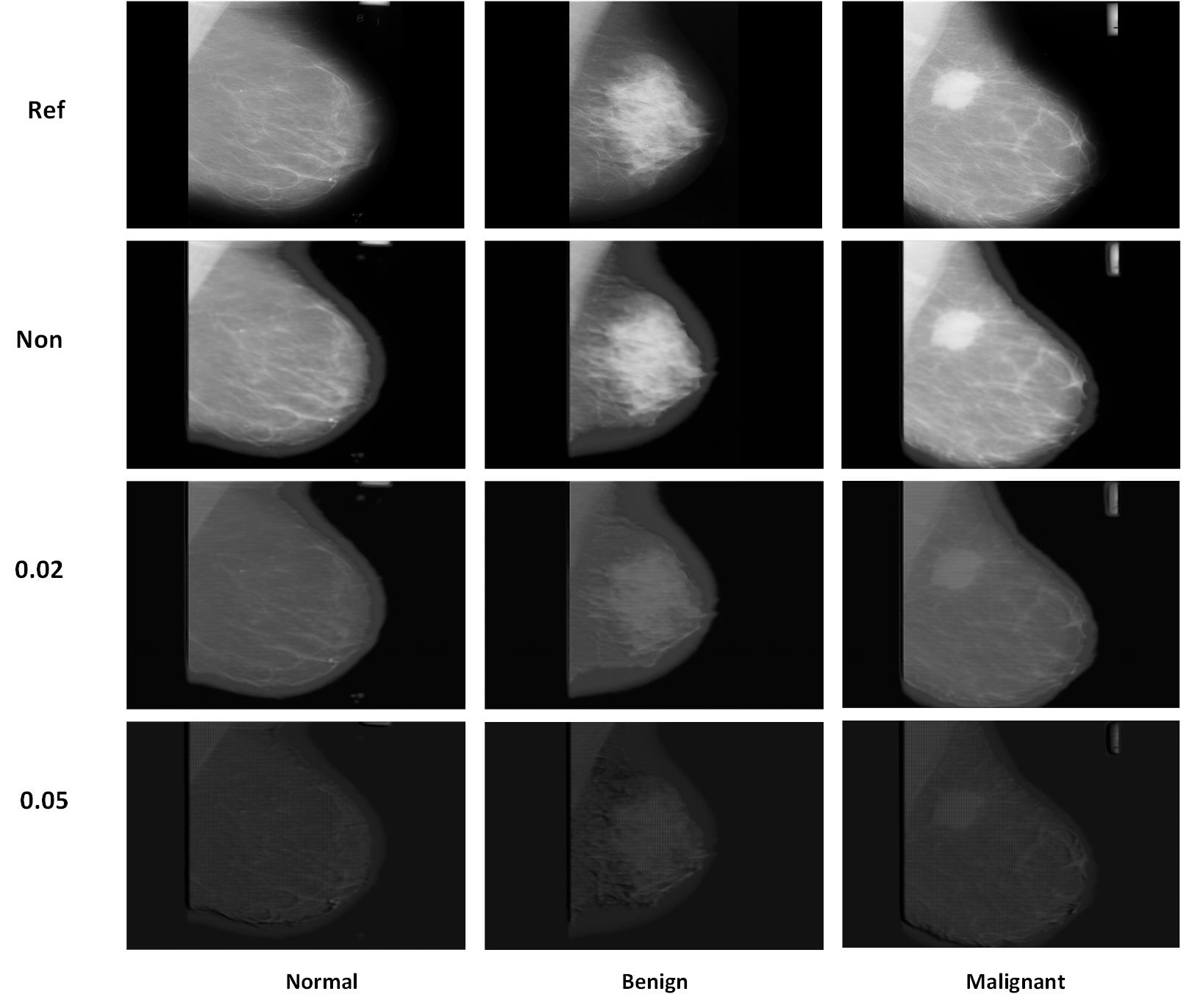}
  \end{center}
  \vspace{-5mm}
    \caption{Recovered MIAS inputs with and without model perturbation.}
      \label{fig:MIA_Perturbation_MIAS}
\end{figure*}

\begin{figure*}[!htb]
  \begin{center}
    \includegraphics[width=0.7\textwidth, height=6.5cm]{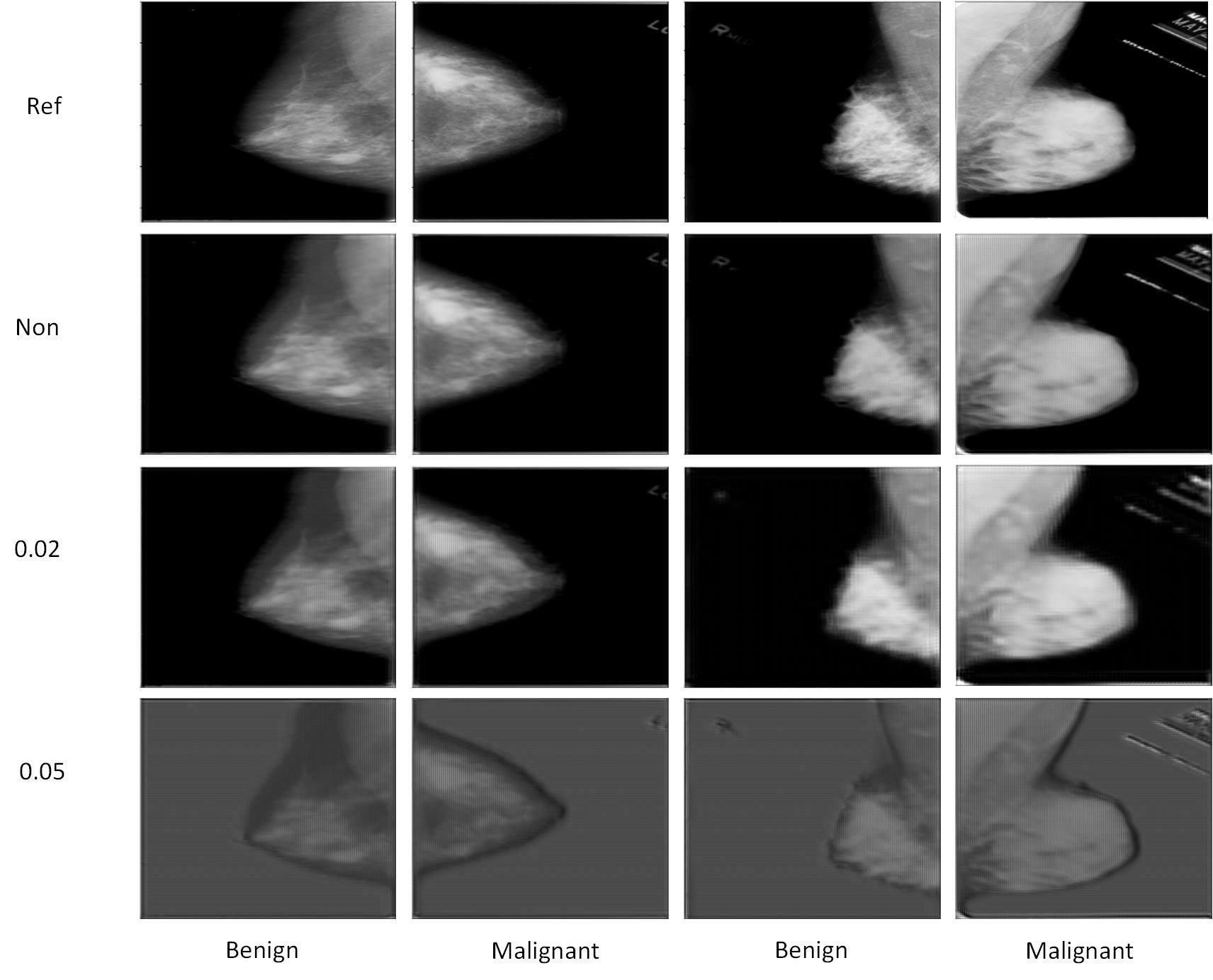}
  \end{center}
  \vspace{-5mm}
    \caption{Recovered CBIS-DDSM inputs with and without model perturbation.}
      \label{fig:MIA_Perturbation_DDSM}
\end{figure*}

Fig. \ref{fig:MIA_Perturbation_MIAS} and Fig. \ref{fig:MIA_Perturbation_DDSM} show the defense performance of model perturbation with different noise scale when the split point is in the 4th layer. We experiment Gaussian noise distributions with scale $0.02$ to $0.05$ and central $0$. When the scale increases, the recovered inputs become more blurry and lose more details.

Table \ref{table2} shows the inference accuracy of CNN models injected by different scale of noises as well as MSE, PSNR and SSIM metrics for the attack. Non-noise means the original trained model without perturbation. When the scale increases, the recovered inputs have higher MSE, PSNR, and lower SSIM. Model perturbation reduces the quality of recovered inputs while slightly decreases the inference performance. These results provide an intuitive guide for medical staff to adjust the scale of noises in defense against the model inversion attack while keeping a satisfied model performance.

\begin{table*}[!htb]
	\renewcommand{\arraystretch}{1}
    \footnotesize
	\caption{MSE, PSNR, SSIM for deep learning model with and without model perturbation}\label{table2} \centering
	\tabcolsep=5pt
	\begin{tabular}{|p{1.2cm}|p{1.2cm}|p{1.2cm}|p{1.2cm}|p{1.2cm}|p{1.2cm}|p{1.2cm}|}
		\hline
		Datasets & \multicolumn{3}{c|}{MIAS} &  \multicolumn{3}{c|}{DDSM}\\
		\hline
		Noise & non & 0.02 & 0.05 & non & 0.02 & 0.05 \\
		\hline
		Accuracy & $0.62$ & $0.60$ & $0.55$ & $0.618$ & $0.582$ & $0.553$\\
		\hline
		MSE & $55.042$ & $1992.121$ & $6163.927$ & $29.649$ & $327.7$ & $4238.92$\\
       	\hline
		PSNR & $31.039$ & $15.479$ & $10.528$ & $33.962$ & $23.072$ & $11.897$\\
       	\hline
		SSIM & $0.994$ & $0.714$ & $0.170$ & $0.995$ & $0.608$ & $0.523$\\
        \hline
	\end{tabular}
\end{table*}


\section{Conclusion}
In this paper, we have evaluated two types of inference attacks on medical images and clinical records, and demonstrated that these attacks can infer sensitive attributes of medical health records as well as recover medical images at high fidelity. Our research finding exposes the risk of privacy leakage for using deep learning models in training medical data. To circumvent this problem, we proposed inference attack defenses based on label perturbation and model perturbation. Experimental results showed that the proposed defenses can effectively defend the malicious inference attacks while the deep learning performance can still be preserved commendably. The experimental results and the approaches presented help to raise awareness about the privacy issues of deploying deep learning networks in medicine and potentially open up a new vista to ensure patients' privacy and confidentiality in the increasing adaptation of AI-enabled information infrastructure in healthcare delivery and medical research.

\end{document}